\documentclass[letterpaper, 10 pt, conference]{ieeeconf}  
\IEEEoverridecommandlockouts

\usepackage[utf8]{inputenc}
\usepackage[T1]{fontenc}
\usepackage{amsmath}
\usepackage{amsfonts}
\usepackage{amssymb}

\usepackage{amsthm}
\theoremstyle{plain}
\usepackage{float}
\usepackage{graphicx}
\usepackage{booktabs}
\usepackage[font=small]{caption}
\usepackage{subcaption}
\usepackage[ruled]{algorithm2e}
\usepackage{algorithmic}
\usepackage{enumerate}
\usepackage{times}
\usepackage{soul}
\usepackage{url}
\usepackage{bm}
\usepackage{tikz}
\usepackage{wrapfig}
\usepackage[table]{xcolor}
\usepackage{colortbl}
\usepackage{adjustbox}
\usepackage{placeins}
\usepackage{array, multirow}
\usepackage{pifont}
\usepackage{cuted}    
\usepackage[nocompress]{cite}
\usepackage{authblk}
\makeatletter
\renewcommand\AB@affilsepx{; \protect\Affilfont}
\makeatother
\makeatletter
\let\NAT@parse\undefined
\makeatother
\usepackage{hyperref}

\hypersetup{
    colorlinks=true,
    urlcolor=blue,
    citecolor=black,
    linkcolor=black,
}
\usepackage[capitalise]{cleveref}
\newcommand{\bb}{\mathbf}

\begin{document}

\newcommand\mycommfont[1]{\footnotesize\rmfamily\textcolor{blue}{#1}}
\usetikzlibrary{arrows.meta}
\usetikzlibrary{positioning}
\tikzstyle{decision} = [diamond, draw, fill=blue!20,
    text width=6em, text badly centered, node distance=3cm, inner sep=0pt]
\tikzstyle{block} = [rectangle, draw, fill=gray!10,
    text width=10em, very thick, text centered, rounded corners, minimum height=2.2em]
\tikzstyle{line} = [draw, -{latex[scale=15.0]}]
\tikzstyle{cloud} = [draw, ellipse,fill=red!20, node distance=3cm,
    minimum height=2em]
\setlength{\fboxrule}{1pt}
\setlength{\fboxsep}{0pt}

\newcounter{task}
\newcommand{\task}[2]{%
  \refstepcounter{task}%
  \label{#1}%
  \textit{Task~\thetask:\ #2}%
}

\newcommand{\red}[1]{\textcolor{black}{#1}}
\newcommand{\green}[1]{\textcolor{green}{#1}}
\newcommand{\blue}[1]{\textcolor{blue}{#1}}

\captionsetup{skip=1pt}
\setlength{\textfloatsep}{1pt}
\setlength{\belowdisplayskip}{1pt} \setlength{\belowdisplayshortskip}{1pt}
\setlength{\abovedisplayskip}{1pt} \setlength{\abovedisplayshortskip}{1pt}
\setlength{\floatsep}{1pt} \setlength{\textfloatsep}{1pt}
\setlength{\intextsep}{1pt}
\setlength{\abovecaptionskip}{1pt}
\setlength{\belowcaptionskip}{1pt}
\captionsetup{belowskip=1pt}

\newcommand{\cmark}{\ding{51}}%
\newcommand{\xmark}{\ding{55}}%

\newtheorem{innercustomthm}{Theorem}
\newenvironment{customthm}[1]
  {\renewcommand\theinnercustomthm{#1}\innercustomthm}
  {\endinnercustomthm}

\newtheorem{innercustomprop}{Proposition}
\newenvironment{customprop}[1]

  {\renewcommand\theinnercustomprop{##1}\innercustomprop}
  {\endinnercustomprop}

\newtheorem{definition}{Definition}[section]
\newtheorem{prop}{Proposition}
\newtheorem{theorem}{Theorem}[section]

\captionsetup[figure]{size=small}

\title{\LARGE \bf Rewind-IL: Online Failure Detection and State Respawning\\ for Imitation Learning
}

\author{Gehan Zheng$^{1}$,
Sanjay Seenivasan$^{1,2}$,
Matthew Johnson-Roberson$^{1}$,
Weiming Zhi$^{1,3,4}$%
\thanks{$^{1}$College of Connected Computing, Vanderbilt University, USA.}
\thanks{$^{2}$School of Computer Science, University of Waterloo, Canada.}
\thanks{$^{3}$School of Computer Science, The University of Sydney, Australia.}
\thanks{$^{4}$Australian Centre for Robotics, The University of Sydney, Australia.}}

\maketitle

\setlength{\stripsep}{-15pt}
\vspace{-2em}
\begin{strip}
\centering
\includegraphics[width=\linewidth]{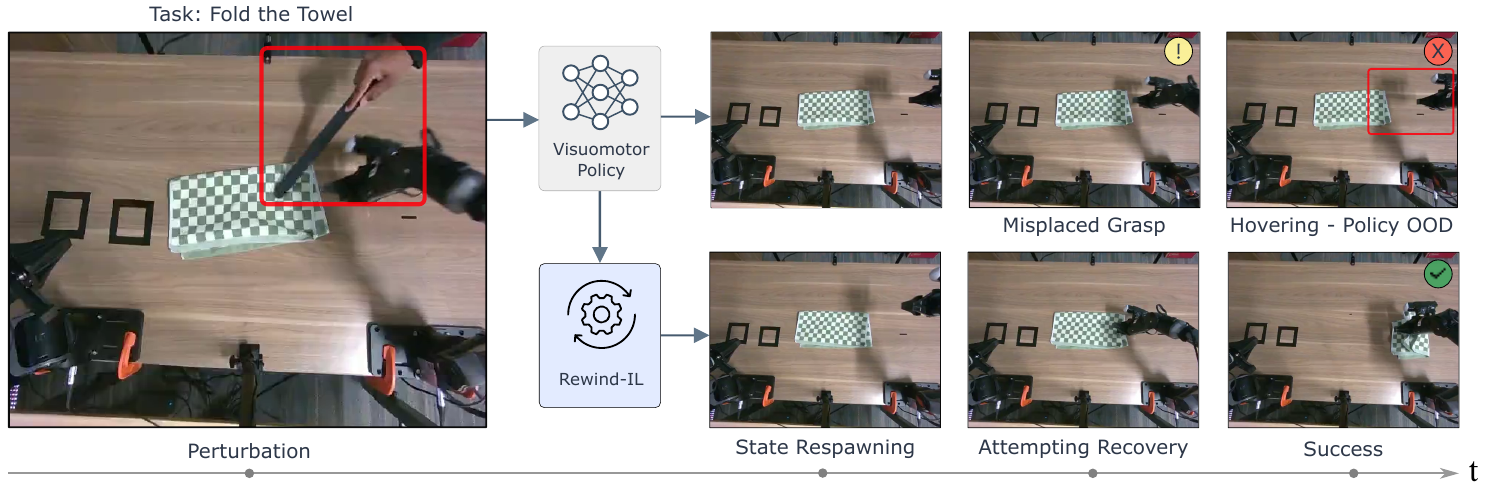} 
\captionof{figure}{\emph{Rewind-IL} enables visuomotor robot policies to efficiently predict and recover from task failures. When the baseline policy fails, the framework rewinds to a prior state to reattempt the task successfully.}
\label{fig:teaser}
\vspace{2em}
\end{strip}

\begin{abstract}
Imitation learning has enabled robots to acquire complex visuomotor manipulation skills from demonstrations, but deployment failures remain a major obstacle, especially for long-horizon action-chunked policies. Once execution drifts off the demonstration manifold, these policies often continue producing locally plausible actions without recovering from the failure. Existing runtime monitors either require failure data, over-trigger under benign feature drift, or stop at failure detection without providing a recovery mechanism. We present Rewind-IL, a training-free online safeguard framework for generative action-chunked imitation policies. Rewind-IL combines a zero-shot failure detector based on Temporal Inter-chunk Discrepancy Estimate (TIDE), calibrated with split conformal prediction, with a state-respawning mechanism that returns the robot to a semantically verified safe intermediate state. Offline, a vision-language model identifies recovery checkpoints in demonstrations, and the frozen policy encoder is used to construct a compact checkpoint feature database. Online, Rewind-IL monitors self-consistency in overlapping action chunks, tracks similarity to the checkpoint library, and, upon failure, rewinds execution to the latest verified safe state before restarting inference from a clean policy state. Experiments on real-world and simulated long-horizon manipulation tasks, including transfer to flow-matching action-chunked policies, demonstrate that policy-internal consistency coupled with semantically grounded respawning offers a practical route to improved reliability in imitation learning.
\end{abstract}


\section{Introduction}

Imitation learning (IL) has become a practical paradigm for teaching robots complex manipulation skills directly from demonstrations, especially where rewards or controllers are hard to specify by hand~\cite{ravichandar_recent_2020}, and recent visuomotor architectures further improve long-horizon performance by predicting extended action chunks rather than single-step controls~\cite{zhao_learning_2023-1,chi_diffusion_2023}. Yet these policies remain brittle at deployment: small execution errors, perception shifts, object motion, or missed contacts push the robot into states poorly covered by the demonstrations, after which the policy often keeps producing locally self-consistent actions while making no task progress. The result is failures such as dropped objects, misaligned placements, or incomplete subtask transitions from which the robot cannot recover.

\begin{figure*}[htbp]
    \centering
    \includegraphics[width=0.95\linewidth]{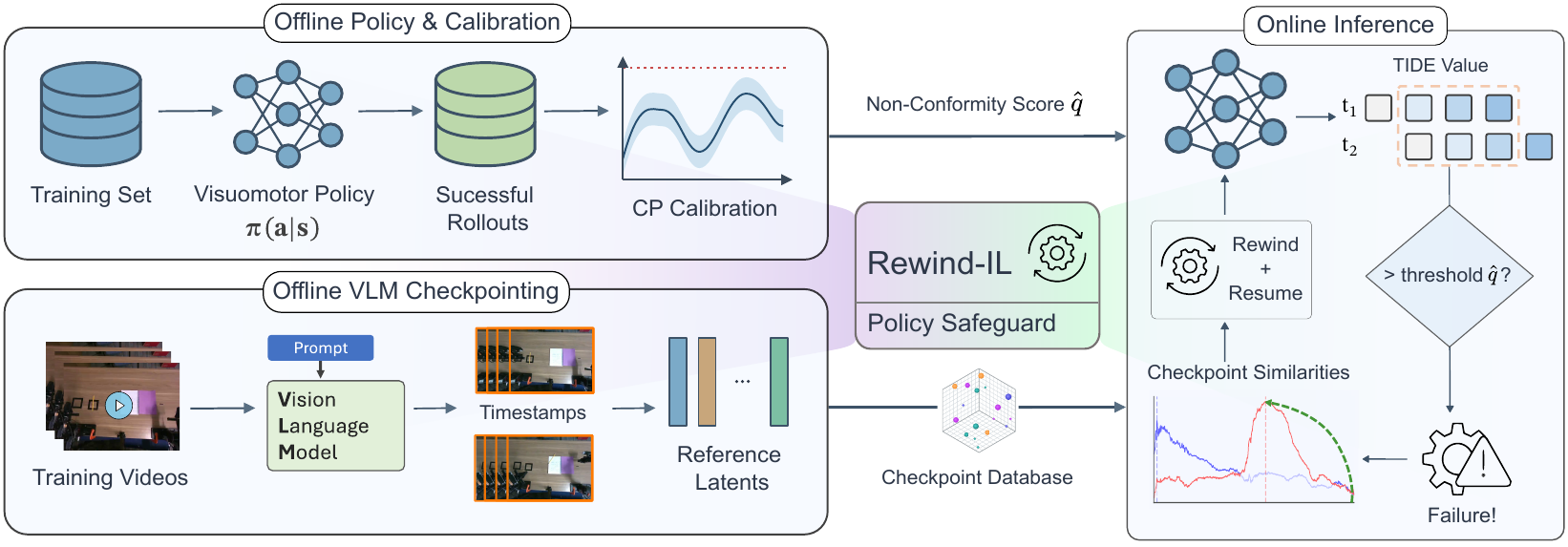}
    \caption{Overview of the \emph{Rewind-IL} framework. \textbf{(Left) Offline Staging:} Successful policy rollouts are used to construct a viable CP threshold for failure detection. Concurrently, a VLM extracts meaningful keyframes from demonstration videos to generate a checkpoint database. \textbf{(Right) Online Policy Deployment:} TIDE flags failures, and the policy returns to a checkpointed state.} 
    \vspace{-2em}
    \label{fig:rewind-il-architecture}
\end{figure*}

Reliable deployment requires answering two questions online: \emph{when} to intervene and \emph{where} to return. Prior runtime monitors flag anomalous states but stop at detection; test-time adaptation methods assume the policy can continue from the current state, forgoing explicit state restoration. We introduce Rewind-IL, a training-free online safeguard that separates detection from recovery-target selection, grounding both in signals inherent to the trained policy and its demonstrations. For detection, \emph{Temporal Inter-chunk Discrepancy Estimate (TIDE)} monitors internal self-consistency---the disagreement between the current action chunk and the plan predicted one step earlier; a sharp rise signals the policy is revising its near-future plan after an unexpected state. TIDE is calibrated by split conformal prediction into a threshold derived solely from successful rollouts.

For recovery, Rewind-IL builds an offline database of semantically verified intermediate states: a VLM identifies recovery timestamps in demonstrations (e.g., completed grasps, subgoal transitions) and the frozen policy encoder extracts compact latent templates. Online, it tracks cosine similarity between the current embedding and each template to identify the most recently confirmed safe state, and upon failure rewinds to it, clears stale policy memory, and restarts inference cleanly. The result is a lightweight safeguard needing no retraining, failure data, or auxiliary controller, built entirely from signals in standard IL pipelines.

We evaluate Rewind-IL on real-world bimanual manipulation tasks and in simulation. TIDE achieves strong failure-detection performance, and the full Rewind-IL framework substantially improves task success both under natural execution failures and adversarial disturbances.

Concretely, the contributions of this paper include:
\begin{enumerate}
\item \emph{Rewind-IL}, an online safeguard framework designed to efficiently detect task failures over generative imitation learning polices based on action chunking;
\item \emph{Temporal Inter-chunk Discrepancy Estimate (TIDE)}, a policy self-consistency signal calibrated by conformal prediction for zero-shot failure detection, together with a VLM-guided checkpoint construction pipeline for selecting semantically meaningful recovery targets;
\item We demonstrate on real-world and simulated long-horizon manipulation tasks that Rewind-IL markedly improves robustness and task success, including under adversarial disturbances.
\end{enumerate}
\section{Related Work}

\noindent\textbf{Failure Detection in Generative and Action-Chunked Policies:}
Action-chunked policies such as Action-Chunking with Transformers (ACT)~\cite{zhao_learning_2023-1} and Diffusion Policy~\cite{chi_diffusion_2023} are prone to compounding errors under distribution shift~\cite{ross_reduction_2011-1}, motivating a line of work on runtime OOD detection.
Architecture-specific approaches extract uncertainty from diffusion denoising~\cite{he_uncertainty_2025,rosasco_kdpe_2025-1} or internal distillation signals~\cite{song_structurally-fused_2025} but are tied to their model families; diffusion OOD detection also extends to $\mathbb{SE}(3)$ pose sequences~\cite{cheng_dose3_2025}, targeting pose- rather than policy-level anomalies.
FIPER~\cite{romer_failure_2025} and FAIL-Detect~\cite{xu_can_2025-1} are more general, combining encoder-level OOD scores with conformal prediction~\cite{angelopoulos_conformal_2023}, yet anchor their anomaly scores to a frozen reference distribution, so benign feature drift can be misclassified as failure even under a time-varying threshold.
Sentinel~\cite{agia_unpacking_2025} partially addresses this by pairing a temporal-consistency detector with online VLM queries.
In contrast, TIDE requires no training and applies across action-chunked architectures: it monitors the self-consistency of overlapping chunks and calibrates thresholds via split conformal prediction, generalizing to flow-matching policies.

\noindent\textbf{Error Recovery and State Backtracking:}
Traditional reactive mechanisms, e.g. CBFs~\cite{ames_control_2019-1}, heuristic recoveries~\cite{wang_infer_2026}, and fixed-pose resets~\cite{santhanam_reliable_2026}, either stop the robot or return it to a hardcoded geometric state, ignoring semantic task context.
CycleVLA~\cite{ma_cyclevla_2026} goes further by using a VLM to detect failure and trigger subtask backtracking with MBR decoding at retry, but incurs costly online inference and backtracks only at coarse subtask granularity.
Rewind-IL instead respawns to a fine-grained, \emph{semantically verified} intermediate state identified entirely offline, enabling recovery at negligible online cost.

\noindent\textbf{Semantic Verification and VLM-Guided Monitoring:}
EVE~\cite{ali_eve_2026}, FOREWARN~\cite{wu_foresight_2025-1}, and CoVer-VLA~\cite{kwok_scaling_2026} integrate VLMs directly into the control loop to verify or score candidate action chunks at every step, achieving strong semantic grounding at the cost of prohibitive inference latency.
Rewind-IL avoids this trade-off by restricting VLM use to an \emph{offline} checkpoint identification phase; online monitoring degrades to a lightweight cosine similarity search over a frozen-encoder feature database, preserving semantic rigor at closed-loop control frequencies.

\section{Preliminaries}\label{sec:preliminaries}

\noindent\textbf{Behavior Cloning and Action Chunking:} We consider standard behavior cloning (BC), which learns a control policy directly from expert demonstrations. Given a dataset of $N$ demonstrations $\mathcal{D} = \{\tau_1, \dots, \tau_N\}$, each trajectory is a sequence of observation--action pairs $\tau_i = \{(o_1, a_1), \dots, (o_{T_i}, a_{T_i})\}$, where $o_t \in \mathcal{O}$ is the observation (images and proprioception) and $a_t \in \mathcal{A}$ the expert action. BC learns a parametric policy $\pi_\theta$ by supervised learning: a single-step policy predicts $\pi_\theta(a_t \mid o_t)$ by minimizing the negative log-likelihood $\mathbb{E}_{(o,a)\sim\mathcal{D}}[-\log \pi_\theta(a \mid o)]$, while an \emph{action-chunked} policy outputs a short horizon of $K$ future actions $\hat{\mathbf{a}}_{t:t+K-1} = \pi_\theta(o_t)$, trained to match expert chunks via $\mathbb{E}[\|\hat{\mathbf{a}}_{t:t+K-1} - \mathbf{a}_{t:t+K-1}\|]$. Chunking improves temporal consistency and captures short-horizon structure over long horizons; ACT~\cite{zhao_learning_2023-1} is representative, and overlapping chunks are typically aggregated at inference for smoother control. BC remains sensitive to distribution shift: small execution or perception errors move the robot off the demonstration distribution, where errors accumulate into unrecoverable failures~\cite{ross_reduction_2011-1}.

\noindent\textbf{Conformal Prediction (CP):} CP is a distribution-free framework that quantifies uncertainty by constructing prediction sets with finite-sample coverage guarantees~\cite{angelopoulos_conformal_2023}. Given non-conformity scores $\mathcal{S}_{cal} = \{s_1,\dots,s_n\}$ on exchangeable calibration samples and a target error rate $\alpha$, split CP sets the threshold
\vspace{2pt}
\begin{equation}
    \hat{q} = {\rm Quantile}\left( \mathcal{S}_{cal}, \frac{\lceil (n + 1)(1 - \alpha) \rceil}{n} \right)     \label{eq:cp_threshold}
\end{equation}
so that the prediction set $\mathcal{C}(X_{n+1}) = \{Y : s(X_{n+1},Y) \le \hat{q}\}$ contains the true label with probability at least $1-\alpha$, i.e.\ $\mathbb{P}(Y_{n+1} \in \mathcal{C}(X_{n+1})) \ge 1-\alpha$.

\section{Rewind-IL}\label{sec:Methods}
\subsection{Overview}
Rewind-IL is a training-free online safeguard framework for generative action-chunked robot policies that provides two capabilities: (1)~zero-shot \emph{real-time failure detection} based on the internal self-consistency of the policy's action predictions, and (2)~\emph{state respawning} that physically returns the robot to a semantically verified safe intermediate state when a failure is flagged. The framework is broadly illustrated in Figure \ref{fig:rewind-il-architecture}. In this paper, we instantiate Rewind-IL with ACT as baseline visuomotor policy; the framework is designed to apply broadly to action-chunked policies.

\vspace{0.2em}\noindent\textbf{Calibration (offline):} We roll out the trained policy to obtain $N$ successful episodes and record per-frame \emph{Temporal Inter-chunk Discrepancy Estimate} (TIDE) values, from which split conformal prediction~\cite{angelopoulos_conformal_2023} derives a failure threshold $\hat{q}$.

\vspace{0.2em}\noindent\textbf{Checkpoint database (offline):} A VLM identifies $K$ semantically meaningful recovery timestamps per training episode; the frozen policy encodes those frames into latent vectors, and a kernel density estimation (KDE) step selects a per-slot embedding into a \emph{checkpoint feature database}.

\noindent\textbf{Inference (online):} At each step we \emph{(i)}~compute TIDE and test it against $\hat{q}$, \emph{(ii)}~measure the cosine similarity of the current latent to every template, and \emph{(iii)}~update per-slot running-maximum bookkeeping gated on the most-similar slot. On a flag with a \emph{peaked} slot available, Rewind-IL rewinds to that checkpoint, clears policy caches, and restarts from a clean state (\cref{alg:rewind}; Figure \ref{fig:teaser}).
\begin{algorithm}[t]
\small
\SetAlgoLined
\DontPrintSemicolon
\KwIn{Policy $\pi_\theta$, templates $\{\hat{\bb{t}}_k\}_{k=1}^K$, threshold $\hat{q}$, peak gap $\Delta_{\mathrm{peak}}$}

Initialize template slot trackers\;
\For{each timestep $t = 0, 1, 2, \ldots$}{
  Observe $o_t$; run $\pi_\theta(o_t) \to (a_t, \tilde{\bb{A}}_t, \bb{f}_t)$\;
  $\text{TIDE}_t \leftarrow \text{compute\_tide}(\tilde{\bb{A}}_t, \hat{\bb{A}}_{t-1})$\;
  \text{update\_slots}$(\bb{f}_t, a_t)$\;
  \eIf{$\mathrm{TIDE}_t > \hat{q}$ \textbf{and} $\exists k:\,\mathrm{peaked}_k$}{
    $k^* \leftarrow$ most recently peaked slot\;
    Rewind to checkpoint $k^*$ (direct return)\;
    Reset ensembler and action queue\;
  }{
    Execute normal action $a_t$\;
  }
}
\caption{Rewind-IL Online Inference Loop}
\label{alg:rewind}
\end{algorithm}

\subsection{Error Flagging via TIDE}

\noindent\textbf{Temporal Inter-chunk Discrepancy Estimate:} A baseline visuomotor policy's temporal ensembler~\cite{zhao_learning_2023-1} maintains an aggregated action plan $\hat{\bb{A}}_{t-1} \in \mathbb{R}^{B \times T \times D}$ assembled from prior inference steps, where $B$ is batch size, $T$ is the overlap length with the new chunk, and $D$ is the action dimension. 

\begin{equation}
    \mathrm{TIDE}_t = \frac{1}{BDT}\sum_{b,d,\tau}
    \bigl(\hat{A}^{(b)}_{t-1,\tau,d} - \tilde{A}^{(b)}_{t,\tau,d}\bigr)^2
    \label{eq:tide2}
\end{equation}

At step $t$ the policy produces a fresh chunk $\tilde{\bb{A}}_t \in \mathbb{R}^{B \times T' \times D}$ with $T' \ge T$. We align the first $T$ steps and compute TIDE as the mean squared discrepancy defined in Eq. \ref{eq:tide2}. TIDE measures how much the policy's current belief about the near future diverges from its one-step-earlier prediction. ACT's temporal ensembler~\cite{zhao_learning_2023-1} assumes overlapping chunks are mutually consistent under nominal execution---the property that makes chunk averaging meaningful; TIDE turns this assumption into an explicit runtime criterion.
\begin{prop}[Temporal consistency]
\label{prop:tide_lipschitz}
Let $\mathcal{M}$ be the manifold of states covered by training demonstrations, and let $o_t$ denote a compact observation (e.g.\ proprioceptive state or encoder latent).
Assume $\pi_\theta$ is locally $L$-Lipschitz on $\mathcal{M}$ (i.e.\ $\|\pi_\theta(o) - \pi_\theta(o')\| \le L\|o - o'\|$ for $o, o' \in \mathcal{M}$) and approximately unimodal on $\mathcal{M}$.
If $o_t \in \mathcal{M}$ and $\|o_t - o_{t-1}\|_2 \le \epsilon$, then $\mathrm{TIDE}_t \lesssim L^2\epsilon^2$.
When $o_t \notin \mathcal{M}$, the local Lipschitz constant may grow sharply or a mode switch may occur, causing $\mathrm{TIDE}_t$ to substantially exceed this bound.
\end{prop}

\noindent Aligning the first $T$ steps makes both tensors predict the same window, so Lipschitzness gives $\|\tilde{\bb{A}}_t[:T]-\hat{\bb{A}}_{t-1}\|\lesssim L\epsilon$ and squaring yields $\mathrm{TIDE}_t\lesssim L^2\epsilon^2$. Locality avoids requiring a global Lipschitz constant; unimodality excludes intent switches (rare within a manipulation phase); the $\lesssim$ absorbs the smoothing of the aggregate $\hat{\bb{A}}_{t-1}$.

\begin{figure}[!t]
    \centering
    \includegraphics[width=\linewidth]{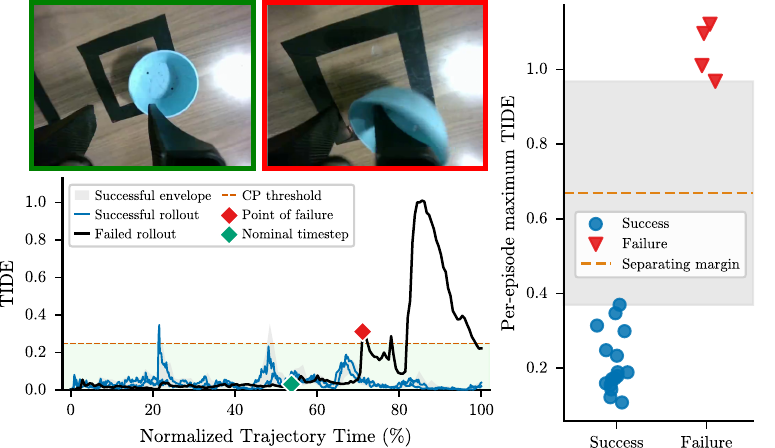}
    \caption{TIDE on \emph{Cup and Box}. \textbf{(a)}~TIDE over normalized progress for successful rollouts (min--max envelope) and a failed rollout, which crosses the CP threshold at failure onset (point of failure, red; nominal timestep, green). \textbf{(b)}~Per-episode maximum TIDE separates successful from failed episodes by a wide margin.}
    \label{fig:tide_evolution}
    \vspace{-0.3em}

\end{figure}

\vspace{0.2em}\noindent\textbf{Failure detection via conformal prediction:} A failure is flagged when:
\begin{equation}
    \mathrm{is\_failing}(t) = \mathbf{1}\!\bigl[\,\mathrm{TIDE}_t > \hat{q}\,\bigr].
    \label{eq:failure_crit2}
\end{equation}
The threshold $\hat{q}$ is set offline by split conformal prediction~\cite{angelopoulos_conformal_2023}, following~\cite{xu_can_2025-1}. Let $\mathcal{S}_{\mathrm{cal}} = \{s_1,\dots,s_n\}$ be the per-frame TIDE values collected on held-out \emph{successful} calibration episodes (trimming $\Delta$ boundary frames per episode to exclude startup and completion artifacts). The threshold is set at the empirical $(1\!-\!\alpha)$-quantile. We use $\alpha = 0.001$ in all experiments. \Cref{fig:tide_evolution}(a) illustrates how TIDE is used to flag an execution failure, contrasting a nominal timestep below the threshold with the first point of failure.

\subsection{Offline Checkpoint Database Construction}

The checkpoint database is built offline in two stages: VLM identification of safe recovery timestamps, then extraction of latent feature templates at those timestamps.

\vspace{0.2em} \noindent\textbf{VLM-guided safe timestamp identification:} For each training episode we upload the full middle-camera video, downsampled to 1~FPS, and query a VLM (Gemini 3.1 Pro, \texttt{temperature}~$=0.1$, structured-JSON output) for timesteps at which the robot reaches an unambiguous, task-consistent intermediate state suitable for recovery (e.g.\ just after a grasp closes or at a sub-goal transition; the final completion state is excluded). The VLM returns an ordered per-episode list of integer-second \emph{slot timestamps} $\bigl\{t_1^{(e)},\dots,t_K^{(e)}\bigr\}$. Because Rewind-IL runs per task on a single-task policy and the VLM reports subtask transitions in consistent temporal order, slot $k$ denotes the same subgoal across episodes. We set $K$ to the modal timestamp count and discard episodes deviating from it, removing missed or spurious detections; across our six tasks this modal-count agreement averages $96\%$ ($92$--$100\%$). The step is fully automatic and needs no human verification, though manual annotation is a cheap alternative ($K$ frames per episode).

\begin{figure*}[t!]
    \centering
    \includegraphics[width=0.95\linewidth]{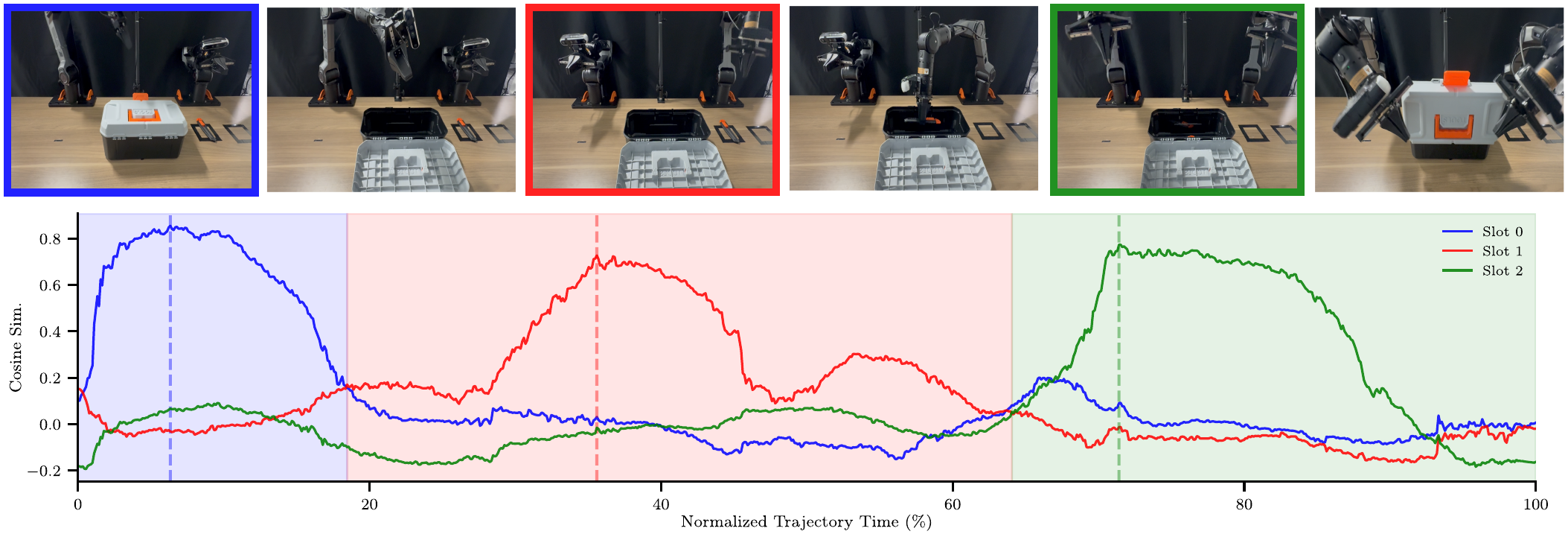}
    \caption{Online similarity tracking on \emph{Toolbox and Knife}: per-slot cosine similarity to reference checkpoint features; the chosen checkpoint per subtask is highlighted above.}
    \label{fig:checkpointing_viz}
    \vspace{-2.2em}
\end{figure*}

\noindent\textbf{Policy feature extraction:} At each VLM-identified checkpoint frame $(e, t_k^{(e)})$ we perform a single forward pass of the frozen policy and capture the observation-conditioned part of the policy encoder's output, which is a token sequence $(f_0, f_1,\dots,f_{L-1}) \in \mathbb{R}^{L \times d}$. We mean-pool the observation-conditioned tokens:
\begin{equation}
    \bb{f}_{\mathrm{enc}} = \frac{1}{L}\sum_{i=0}^{L-1} f_i \;\in \mathbb{R}^{d}.
    \label{eq:enc_feat}
\end{equation}
This $d$-dimensional vector summarizes the policy's interpretation of the current observation.

\vspace{0.2em} \noindent\textbf{KDE-based representative template selection:} For slot $k$, the cross-episode feature vectors $\{\bb{f}_k^{(e)}\}_{e=1}^{E}$ form a point cloud in $\mathbb{R}^d$. A simple mean $\bar{\bb{f}}_k = \frac{1}{E}\sum_e \bb{f}_k^{(e)}$ blurs distinct execution trajectories and may not correspond to any real robot state; instead we select the single most \emph{representative} episode embedding via kernel density estimation. As the features are high-dimensional, we first apply PCA retaining a fraction $r = 0.95$ of the variance, projecting to $d_{\mathrm{pca}} \ll d$ dimensions, then fit a Gaussian KDE with the Silverman bandwidth~\cite{silverman_density_2018}:
\begin{equation}
    h = \bar{\sigma}\cdot E^{-1/(d_{\mathrm{pca}}+4)},
    \qquad
    \bar{\sigma} = \frac{1}{d_{\mathrm{pca}}}\sum_{j=1}^{d_{\mathrm{pca}}} \sigma_j,
    \label{eq:silverman}
\end{equation}
where $\sigma_j$ is the empirical standard deviation of the $j$-th PCA dimension. The template for slot $k$ is the \emph{original}-space feature of the episode with the highest estimated log-density in PCA space:
\begin{equation}
    \bb{t}_k = \bb{f}_k^{(e^*)},
    \quad
    e^* = \arg\max_{e}\;\hat{p}\,\!\bigl(\mathrm{PCA}(\bb{f}_k^{(e)})\bigr).
    \label{eq:kde_select}
\end{equation}
Geometrically, $e^*$ is the checkpoint closest to the mode of the demonstration cluster---the most \emph{typical} example rather than a possibly non-existent centroid---mirroring mean-shift clustering~\cite{comaniciu_mean_2002} and yielding sharper matching targets than a mean template. This matters most under multimodal demonstrations: on the multimodal Close Fridge task (\cref{sec:EXPERIMENT}), replacing the KDE template by the mean drops end-to-end success from $70\%$ to $62\%$, as the mean of two modes need not be executable. All slot templates are stored in the checkpoint database for live matching.

\subsection{Online Similarity Tracking and State Respawning}

During inference, Rewind-IL runs two parallel processes: a per-slot similarity tracker monitoring how close the current state is to each checkpoint, and a recovery executor triggered when TIDE fires.

\vspace{0.2em} \noindent\textbf{Per-slot cosine similarity:} At each step a forward hook captures the policy's intermediate feature $\bb{f}_t$ (same extraction as \cref{eq:enc_feat}). The cosine similarity to each pre-normalised template is a single batched dot product:
\begin{equation}
    s_k(t) = \hat{\bb{f}}_t \cdot \hat{\bb{t}}_k, \qquad k = 1,\dots,K,
    \label{eq:cosine_sim}
\end{equation}
where $\hat{\bb{f}}_t = \bb{f}_t / \|\bb{f}_t\|_2$. Stacking the pre-normalised templates as rows of $\hat{\bb{T}} \in \mathbb{R}^{K \times d}$ reduces this to one matrix-vector product $\bb{s}(t) = \hat{\bb{T}}\,\hat{\bb{f}}_t$, efficient even for large $K$.

\vspace{0.2em} \noindent\textbf{Peak detection and action snapshotting:} Each slot $k$ maintains a running maximum similarity and the action snapshot at which it was achieved:
\begin{equation}
\begin{split}
    s_k^{\max}(t) &= \max_{t' \le t} s_k(t'), \\
    a_k^*(t) = a_{t^*_k}, &\quad t^*_k = \arg\max_{t' \le t} s_k(t').
\end{split}
\label{eq:slot_update}
\end{equation}
The running maximum keeps updating whenever a higher similarity is later observed, even after a slot has been flagged; ``peaking'' is therefore a flag, not a separate extremum. Slot $k$ is declared \emph{peaked} once its similarity has failed to improve for more than $\Delta_{\mathrm{peak}}$ consecutive steps:
\begin{equation}
    \mathrm{peaked}_k(t) = \mathbf{1}\!\bigl[t - t^*_k > \Delta_{\mathrm{peak}}\bigr].
    \label{eq:peak_detect}
\end{equation}
Peaking alone does not fix the respawn target. Let $b(t) = \arg\max_k s_k(t)$ be the slot most similar to the current state. The active target $k^*$ is updated only when the currently most-similar slot has itself peaked and differs from the current target:
\begin{equation}
    k^* \leftarrow b(t) \quad\text{if } t - t^*_{b(t)} > \Delta_{\mathrm{peak}} \text{ and } b(t) \neq k^*,
    \label{eq:target_update}
\end{equation}
and on each switch the previously active slot is \emph{released} ($s_k^{\max}\!\leftarrow\!-\infty$) so it can peak again. This gating is what enables recovery to an \emph{earlier} checkpoint: if a disturbance rolls the scene back, the earlier slot again becomes most similar, re-peaks, and is re-selected. A peaked target thus means the robot has already passed (near) the corresponding VLM-verified safe state and is now moving away. \cref{fig:checkpointing_viz} shows these peaks across an episode.

\vspace{0.2em} \noindent\textbf{Recovery and respawning:} When $\mathrm{is\_failing}(t) = 1$ and a peaked target $k^*$ is available, Rewind-IL \emph{rewinds} to checkpoint $k^*$: the controller returns the robot along a direct trajectory to the recorded checkpoint state; because actions are absolute joint-position commands, the return target is a fixed configuration, reached regardless of where the failure is detected. In our tasks the workspace between failure state and checkpoint is unobstructed, so the direct return is safe; the return step is modular and can be replaced by a conservative reverse replay of the stored actions (retracing the just-executed, known-traversable trajectory) or a motion planner in cluttered workspaces.
On arrival we clear the temporal-ensembler buffer and action queue, invalidating the stale plan, and resume from a clean state. Since $k^*$ adapts to the current scene (\cref{eq:target_update}), repeated failures respawn to the most-appropriate checkpoint, not a fixed one.

This separates \emph{when} to recover (the CP-calibrated TIDE threshold) from \emph{where} (the VLM-verified feature library), and acts as an \emph{implicit manifold projection}: respawning returns the system to a point on $\mathcal{M}$, where the locally Lipschitz policy resumes reliable execution.

\definecolor{ourgray}{gray}{0.92}
\definecolor{badred}{RGB}{255, 230, 230}

\begin{figure*}[t]
    \centering
    \begin{minipage}{0.63\textwidth}
        \centering
        \captionof{table}{Failure detection on the real ACT and flow-matching (FM) tasks (TPR / TNR / Balanced Accuracy, all $\uparrow$). Highlighted columns: TIDE.}
        \label{tab:failure_detection_full}
        \small
        \renewcommand{\arraystretch}{1.3}
        \setlength{\tabcolsep}{3pt}
        \adjustbox{max width=\textwidth}{
            \begin{tabular}{l l ccc ccc ccc ccc >{\columncolor{ourgray}}c >{\columncolor{ourgray}}c >{\columncolor{ourgray}}c}
            \toprule
            \multirow{2}{*}{\textbf{Task}} & \multirow{2}{*}{\textbf{Policy}} &
            \multicolumn{3}{c}{\textbf{FAIL-Detect~\cite{xu_can_2025-1}}} &
            \multicolumn{3}{c}{\textbf{RND~\cite{he_rediffuser_2024}}} &
            \multicolumn{3}{c}{\textbf{Clustering OOD}} &
            \multicolumn{3}{c}{\textbf{Mahalanobis}} &
            \multicolumn{3}{c}{\textbf{TIDE (Ours)}} \\
            \cmidrule(lr){3-5} \cmidrule(lr){6-8} \cmidrule(lr){9-11} \cmidrule(lr){12-14} \cmidrule(lr){15-17}
            & & TPR & TNR & Acc.& TPR & TNR & Acc. & TPR & TNR & Acc. & TPR & TNR & Acc. & TPR & TNR & \textbf{Acc.} \\
            \midrule
            \multirow{2}{*}{Cup and Box} & ACT & 1.00 & 0.57 & 0.79 & 1.00 & 0.93 & 0.96 & 0.75 & 0.50 & 0.63 & 1.00 & 0.69 & 0.84 & 1.00 & 1.00 & \textbf{1.00} \\
             & FM & 1.00 & 0.93 & 0.97 & 1.00 & 1.00 & \textbf{1.00} & 0.25 & 0.88 & 0.56 & 1.00 & 0.63 & 0.81 & 1.00 & 1.00 & \textbf{1.00} \\
            \midrule
            \multirow{2}{*}{Pencil and Notebook} & ACT & 1.00 & 1.00 & \textbf{1.00} & 1.00 & 0.07 & 0.54 & 1.00 & 0.00 & 0.50 & 1.00 & 0.21 & 0.61 & 1.00 & 1.00 & \textbf{1.00} \\
             & FM & 1.00 & 0.93 & \textbf{0.97} & 1.00 & 0.93 & \textbf{0.97} & 1.00 & 0.13 & 0.57 & 1.00 & 0.53 & 0.77 & 1.00 & 0.93 & \textbf{0.97} \\
            \midrule
            \multirow{2}{*}{Box and Wrench} & ACT & 0.83 & 0.64 & 0.74 & 0.83 & 0.93 & 0.88 & 1.00 & 0.50 & 0.75 & 0.83 & 0.07 & 0.45 & 1.00 & 0.93 & \textbf{0.96} \\
             & FM & 0.50 & 1.00 & 0.75 & 0.50 & 0.93 & 0.71 & 0.83 & 0.93 & 0.88 & 0.50 & 0.71 & 0.61 & 1.00 & 0.93 & \textbf{0.96} \\
            \midrule
            \multirow{2}{*}{Drawers and Hammer} & ACT & 0.78 & 0.64 & 0.70 & 0.89 & 0.36 & 0.63 & 0.17 & 0.93 & 0.55 & 0.33 & 0.86 & 0.60 & 1.00 & 1.00 & \textbf{1.00} \\             & FM & 0.67 & 0.64 & 0.65 & 0.78 & 0.27 & 0.53 & 1.00 & 0.45 & 0.73 & 1.00 & 0.27 & 0.64 & 1.00 & 1.00 & \textbf{1.00} \\
            \midrule
            \multirow{2}{*}{Toolbox and Knife} & ACT & 0.83 & 1.00 & 0.92 & 1.00 & 0.93 & 0.96 & 0.67 & 0.36 & 0.51 & 0.67 & 0.29 & 0.48 & 1.00 & 1.00 & \textbf{1.00} \\
             & FM & 0.75 & 0.83 & 0.79 & 1.00 & 0.83 & 0.92 & 0.13 & 0.33 & 0.23 & 0.88 & 0.33 & 0.60 & 1.00 & 1.00 & \textbf{1.00} \\
            \midrule
            \multirow{2}{*}{Folding Towel} & ACT & 1.00 & 0.08 & 0.54 & 0.75 & 0.33 & 0.54 & 0.75 & 0.50 & 0.63 & 1.00 & 0.00 & 0.50 & 0.50 & 0.92 & \textbf{0.71} \\
             & FM & 0.88 & 0.42 & 0.65 & 1.00 & 0.33 & 0.67 & 0.50 & 0.92 & 0.71 & 1.00 & 0.42 & 0.71 & 0.50 & 1.00 & \textbf{0.75} \\
            \midrule
            \multirow{2}{*}{\textbf{Average}} & ACT & 0.91 & 0.66 & 0.78 & 0.91 & 0.59 & 0.75 & 0.72 & 0.47 & 0.60 & 0.81 & 0.35 & 0.58 & 0.92 & 0.98 & \textbf{0.95} \\
             & FM & 0.80 & 0.79 & 0.80 & 0.88 & 0.72 & 0.80 & 0.62 & 0.61 & 0.61 & 0.90 & 0.48 & 0.69 & 0.92 & 0.98 & \textbf{0.95} \\
            \bottomrule
            \end{tabular}
        }
    \end{minipage}
    \hfill
    \begin{minipage}{0.36\textwidth}
        \centering
        \captionof{table}{Real-world task success (\%, 20 rollouts).
Each column pairs a failure detector with a recovery strategy:
\textbf{Ours} = TIDE + checkpoint rewind; \textbf{FD+Ck} = FAIL-Detect +
checkpoint rewind; \textbf{TIDE+Sig} = TIDE + signal-based rewind;
\textbf{FM} = flow-matching variant of ACT (base and with Rewind-IL).}

        \label{tab:success_real}
        \small
        \setlength{\tabcolsep}{4pt}
        \renewcommand{\arraystretch}{1}
        \begin{adjustbox}{max width=\linewidth}
        \begin{tabular}{l c cccc cc}
        \toprule
        \multirow{2}{*}{\textbf{Task}} & \multirow{2}{*}{\textbf{Perturb}} & \multicolumn{4}{c}{\textbf{ACT}} & \multicolumn{2}{c}{\textbf{FM}} \\
        \cmidrule(lr){3-6}\cmidrule(lr){7-8}
        & & Base & Ours & FD+Ck & TIDE+Sig & Base & Ours \\
        \midrule
        \multirow{2}{*}{Cup and Box} & $\times$ & 80 & \textbf{90} & 80 & 80 & 75 & \textbf{90} \\ & $\checkmark$ & 15 & \textbf{85} & 75 & 20 & 20 & \textbf{90} \\ \midrule
        \multirow{2}{*}{Pencil and Notebook} & $\times$ & 70 & \textbf{85} & 80 & 75 & 70 & \textbf{90} \\ & $\checkmark$ & 25 & \textbf{80} & \textbf{80} & 20 & 30 & \textbf{65} \\ \midrule
        \multirow{2}{*}{Box and Wrench} & $\times$ & 65 & \textbf{75} & 70 & 65 & 70 & \textbf{80} \\ & $\checkmark$ & 15 & \textbf{80} & 65 & 20 & 20 & \textbf{75} \\ \midrule
        \multirow{2}{*}{Drawers and Hammer} & $\times$ & 55 & \textbf{75} & \textbf{75} & 50 & 55 & \textbf{70} \\ & $\checkmark$ & 20 & \textbf{75} & 70 & 25 & 25 & \textbf{75} \\ \midrule
        \multirow{2}{*}{Toolbox and Knife} & $\times$ & 70 & \textbf{90} & 80 & 70 & 65 & \textbf{75} \\ & $\checkmark$ & 15 & \textbf{85} & 80 & 15 & 10 & \textbf{70} \\ \midrule
        \multirow{2}{*}{Folding Towel} & $\times$ & 60 & \textbf{65} & 60 & 60 & 60 & \textbf{65} \\ & $\checkmark$ & 20 & \textbf{55} & 35 & 25 & 15 & \textbf{55} \\ \midrule
        \multirow{2}{*}{\textbf{Average}} & $\times$ & 66.7 & \textbf{80.0} & 74.2 & 66.7 & 65.8 & \textbf{78.3} \\ & $\checkmark$ & 18.3 & \textbf{76.7} & 67.5 & 20.8 & 20.0 & \textbf{71.7} \\
        \bottomrule
        \end{tabular}
        \end{adjustbox}
    \end{minipage}
    \vspace{-1em}
\end{figure*}

\section{Empirical Evaluations}\label{sec:EXPERIMENT}

\begin{wrapfigure}{r}{0.55\columnwidth}
\centering
\includegraphics[width=\linewidth]{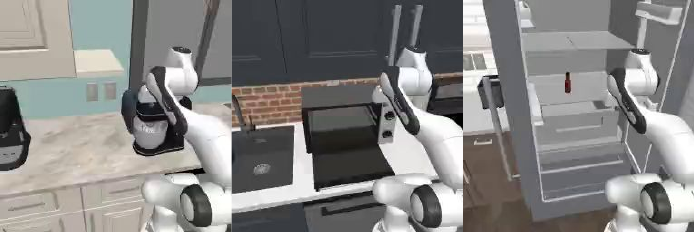}
\caption{RoboCasa Simulation Tasks.}
\label{fig:robocasa}
\end{wrapfigure}

We evaluate Rewind-IL across six bimanual manipulation tasks on a real dual-arm robot and three tasks in the RoboCasa simulation environment \cite{nasiriany_robocasa365_2025} (as depicted in Fig. \ref{fig:robocasa}) with ACT \cite{zhao_learning_2023-1} as the baseline visuomotor action-chunked policy. Videos of all tasks are provided in the supplementary video.
Our evaluation is organized around three central inquiries:
\begin{enumerate}
  \item \textbf{Detection quality.} Does TIDE reliably distinguish genuine task failures from normal execution, compared to other OOD methods?
  \item \textbf{Recovery effectiveness.} Does Rewind-IL's checkpoint-respawning loop improve task success rates under both natural policy failures and adversarial disturbances?
  \item \textbf{Flow-Matching Policies}: Does the framework transfer to flow-matching action-chunked policies?
\end{enumerate}
\begin{figure*}[t!]
    \centering
    \begin{minipage}[t]{0.483\textwidth}
        \centering
        \includegraphics[width=\linewidth]{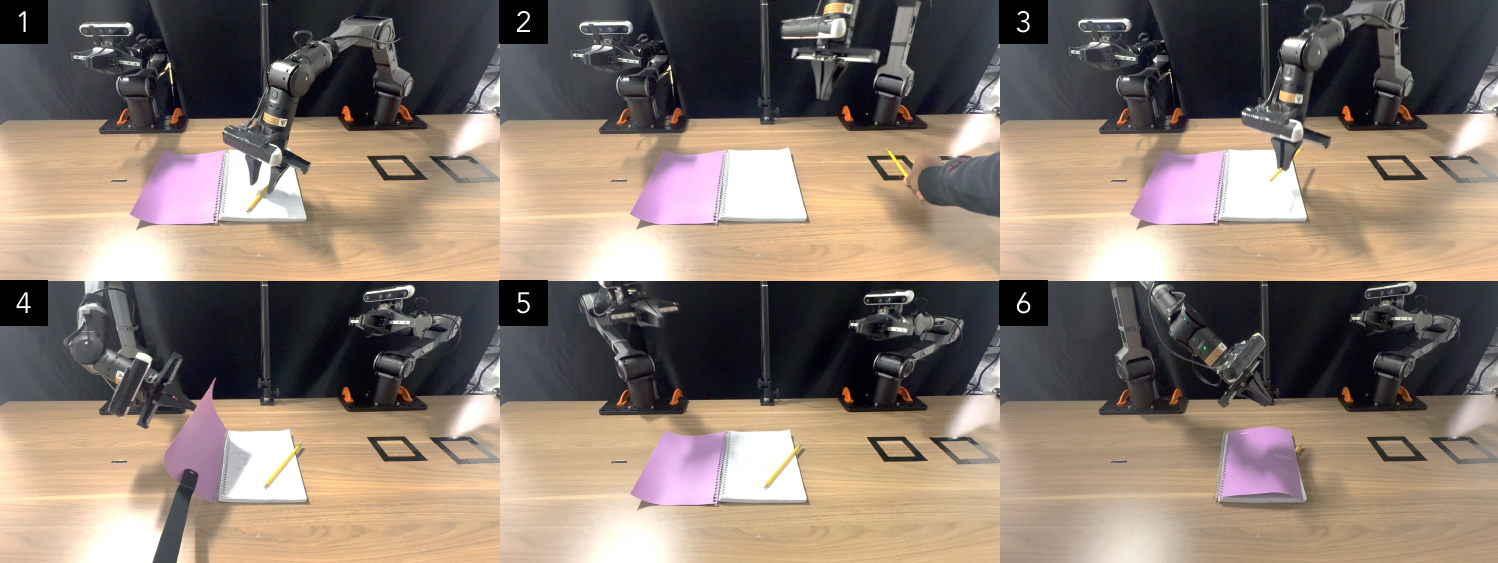}
    \end{minipage}
    \hfill
    \begin{minipage}[t]{0.483\textwidth}
        \centering
        \includegraphics[width=\linewidth]{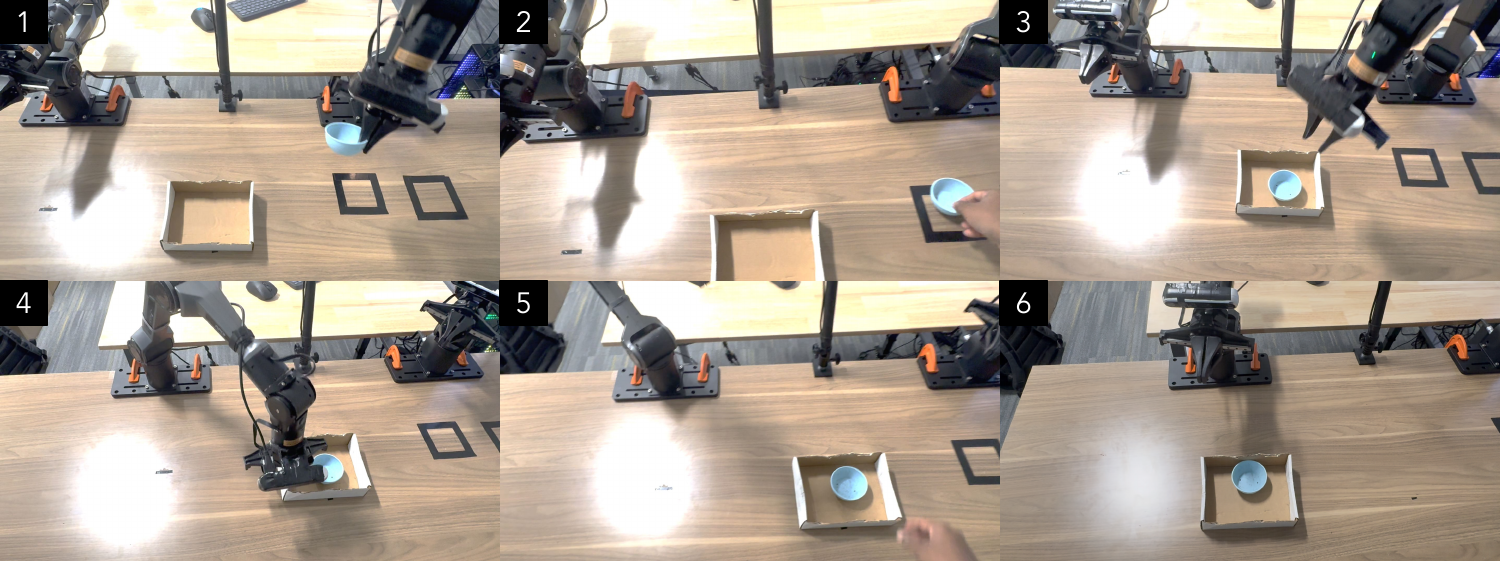}
    \end{minipage}
    \caption{Rewind-IL with perturbations and recovery on \textit{Pencil and Notebook} (bookmarking the page with a pencil and flipping over the front cover) and \textit{Cup and Box} (placing a cup into a tray and moving both to the side of the table by grasping the tray).}
    \label{fig:camera_roll_pencil_cup}
    \vspace{-2em}
\end{figure*}

\subsection{Setup and Metrics}

\noindent\textbf{Real-world setup:} We evaluate on a bimanual system of two AgileX Piper 6-DoF arms with parallel grippers at 30~Hz, across six tasks spanning multi-step coordination (Cup \& Box, Pencil \& Notebook, Box \& Wrench, Drawers \& Hammer, Toolbox \& Knife) and deformable manipulation (Folding Towel). Base ACT models are trained on 50 demonstrations for 100k iterations on NVIDIA RTX 5090 GPUs. All use Force-Aware ACT~\cite{li_tripilot-ff_2026, watanabe_ftact_2025}, adding joint torque $\boldsymbol{\tau}_t \in \mathbb{R}^n$ as proprioceptive input, $\hat{\mathbf{a}}_{t:t+K} = \pi_\theta(\mathbf{I}_t, \mathbf{j}_t, \boldsymbol{\tau}_t)$, which improves bimanual performance.

\noindent\textbf{Evaluation Protocols:}
\begin{itemize}
  \item \textit{Detection.} Evaluated on unperturbed rollouts labelled post-hoc. An episode is labelled predicted-failure if its score crosses the calibrated threshold at \emph{any} timestep (i.e.\ its per-episode maximum exceeds the threshold), the first crossing defining the detection time. The threshold is the $(1-\alpha)$-quantile of the pooled per-frame scores over 10 disjoint successful calibration episodes (no failure data). TPR and TNR are the fractions of failed and successful episodes correctly labelled. All methods are compared on the same rollouts under identical initial layouts and the same disturbance protocol, with thresholds calibrated on a disjoint set of successful episodes, so run-to-run trajectory variability is shared across methods rather than confounding the comparison.
  \item \textit{Recovery (+\,Perturb).} An operator adversarially intervenes during every rollout (e.g., nudging objects, re-opening closed drawers) until a sub-task visibly fails, then stops, directly analogous to the disturbance protocol of MoE-DP~\cite{cheng_moe-dp_2025}, where objects are reset after successful grasps to force re-attempt.
\end{itemize}

\vspace{0.2em} \noindent\textbf{Baselines for failure detection:} We compare TIDE against four post-hoc baselines that utilize successful calibration rollouts to detect OOD states:
\begin{itemize}
    \item \textbf{FAIL-Detect}~\cite{xu_can_2025-1}: A recent SOTA sequential detector that distills policy inputs/outputs into scalar signals and flags failures via conformal prediction.
    \item \textbf{RND / FIPER}~\cite{he_rediffuser_2024, romer_failure_2025}: Random Network Distillation on the policy's encoder embeddings, corresponding to FIPER's RND-OE component~\cite{romer_failure_2025}. FIPER's other component, action-chunk entropy (ACE), requires sampling multiple action chunks per observation, inapplicable to the deterministic latent-mean inference of CVAE-based ACT, which yields a single chunk.
    \item \textbf{Clustering-Based OOD}:  The OOD score is the minimum Euclidean distance to 64 K-means centroids on PCA-project ACT calibration embeddings~\cite{xu_can_2025-1}.
    \item \textbf{Mahalanobis Similarity}: Fits a single Gaussian to the PCA-projected calibration embeddings and reports the Mahalanobis distance~\cite{agia_unpacking_2025}.
\end{itemize}

The embedding-distance baselines (RND, Clustering, Mahalanobis) use a scalar split-CP threshold (\cref{sec:preliminaries}); FAIL-Detect uses an RND-based distillation score with its native time-varying (functional) conformal band; all detectors use $\alpha = 0.001$. Unless noted, Rewind-IL uses $K$ equal to the modal VLM checkpoint count, peak gap $\Delta_{\mathrm{peak}} = 30$ steps, boundary trim $\Delta = 30$ frames, 10 calibration episodes, and the mean-pooled encoder feature ($d = 512$). We report Balanced Accuracy $= (\mathrm{TPR}+\mathrm{TNR})/2$ for detection and task completion rate for recovery. The conservative choice $\alpha = 0.001$ is deliberate: sweeping $\alpha \in \{0.05, 0.01, 0.005, 0.001\}$, TIDE's balanced accuracy improves monotonically as $\alpha$ decreases (average $0.51 \to 0.95$), because a permissive threshold flags benign TIDE spikes at subtask transitions; the calibrated threshold and detection results are invariant across $\Delta \in \{0, 15, 30, 60\}$.

\subsection{Evaluating TIDE as a Failure Detector}

\noindent\textbf{Performance comparison:} TIDE attains the best balanced accuracy on every task, averaging $0.95$ versus $0.78$ for the strongest baseline (FAIL-Detect). \Cref{fig:tide_evolution} shows the TIDE score over a rollout: successful episodes stay within a low nominal band while the failed rollout rises sharply at failure onset, the separation the calibrated detector exploits.

\vspace{0.2em} \noindent\textbf{Why baselines fail:} The embedding-based baselines (Clustering OOD, Mahalanobis) over-trigger severely, misclassifying natural embedding drift as failures (TNR collapsing to $0.00$ on some tasks). RND and FAIL-Detect also rely on fixed reference distributions: RND suffers severe false positives on specific tasks, and while FAIL-Detect is the most robust baseline, it still over-triggers on safe but novel variations, most sharply on deformable Folding Towel. Unlike these methods, whose scores measure distance from a frozen reference distribution, TIDE monitors \emph{intra-policy self-consistency}: by tracking the agreement of the policy's own action-chunk predictions, it is far more robust to benign representation drift, maintaining strong performance across all tasks and largely avoiding the over-triggering that breaks the baselines even on Folding Towel.

\subsection{Task Success Rate: Ablation and Main Results}

Detection and respawning cannot be ablated as fully independent recovery components: respawning requires a detection trigger, and detection used \emph{alone} as a stopping rule would merely terminate the episode early. Importantly, the ACT baseline runs the full horizon with the same chance to self-recover, and is \emph{not} truncated when TIDE fires. The detection ablation is reported in \cref{tab:failure_detection_full}; Tables~\ref{tab:success_real}--\ref{tab:success_sim} measure the integrated benefit over this full-horizon baseline. A finer factorization of detection versus recovery is reported in the FD+Ck and TIDE+Sig columns of \cref{tab:success_real}. 

\begin{figure}[t]
        \centering
        \includegraphics[width=\linewidth]{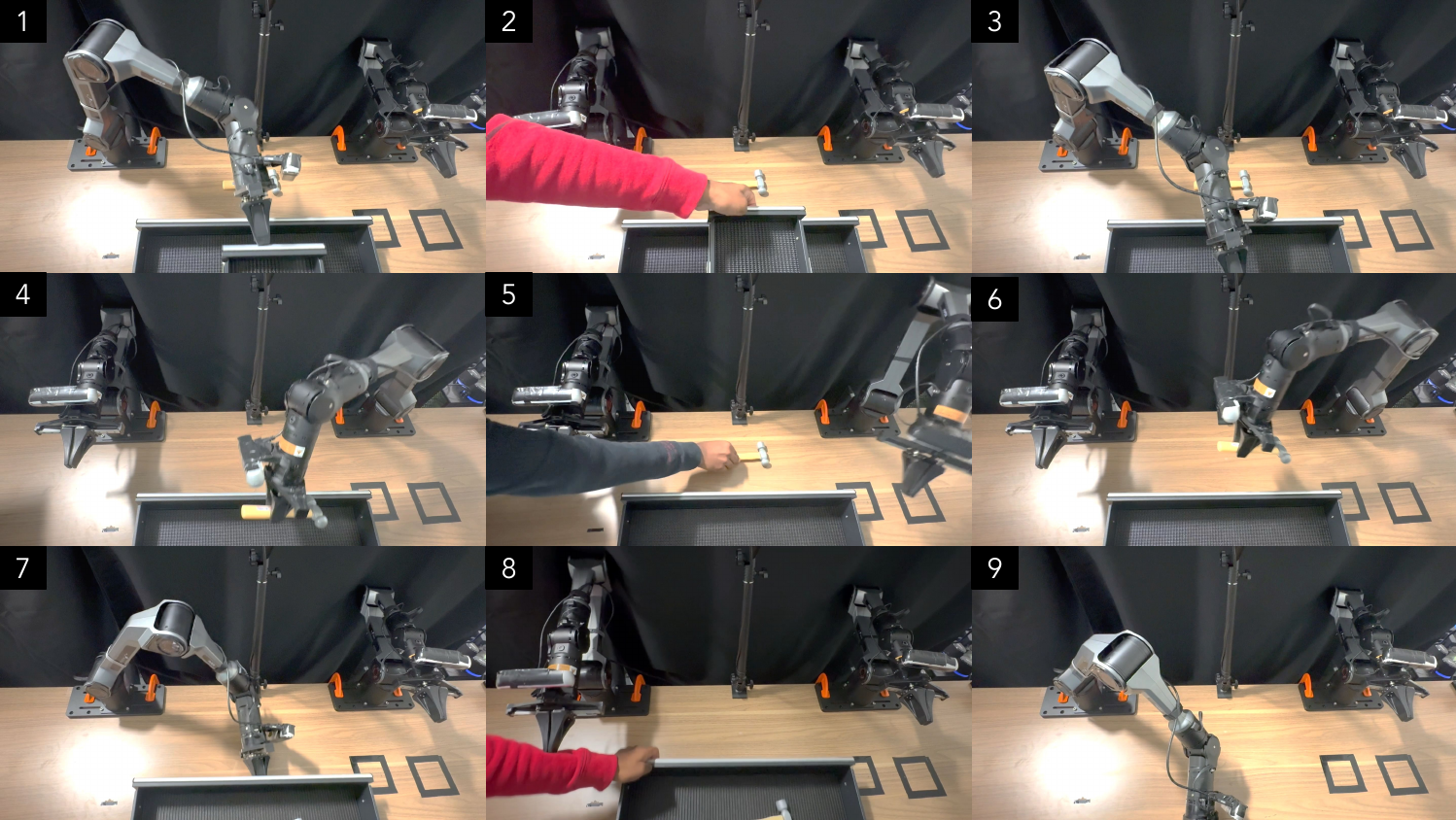}
        \includegraphics[width=\linewidth]{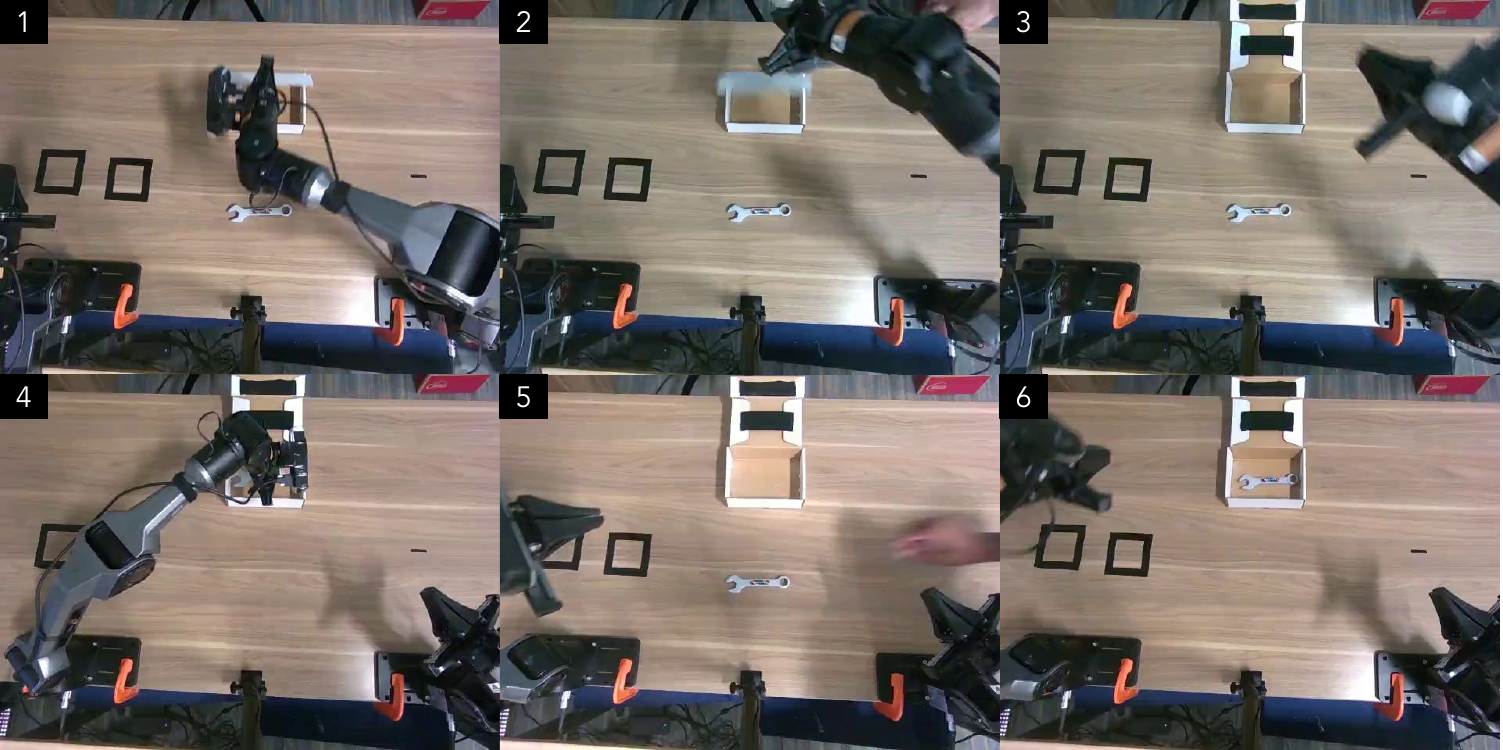}
    \caption{Rewind-IL recovery on \textbf{Drawers and Hammer} (place a hammer inside and close the drawers) and \textbf{Box and Wrench} (lift the box lid and place the wrench inside).}
    \label{fig:camera_roll_drawers_box}
\end{figure}

\noindent\textbf{Effect of Rewind-IL:} ACT\,+\,Rewind-IL consistently outperforms the plain ACT baseline across all nine tasks (Tables~\ref{tab:success_real}--\ref{tab:success_sim}), by $+5$ to $+20$ percentage points~(pp) in the real world and $+15$ to $+20$\,pp in simulation. The largest gains occur on \textit{Drawers\,\&\,Hammer} and \textit{Toolbox\,\&\,Knife}, which require precise multi-step grasps where a single slip compounds into irrecoverable failure---exactly where checkpoint respawning helps most (\cref{fig:camera_roll_drawers_box}). The consistent lift in simulation shows the benefit is not an artifact of our hardware.

\vspace{0.02em} \noindent \textbf{Robustness under adversarial disturbance:} The ``Perturb`` conditions apply the disturbance protocol to \emph{every} rollout, a worst-case stress test. Without recovery, ACT\,+\,Perturb\ rarely completes the task; coupling disturbance with Rewind-IL recovers most of this loss, five of six tasks approaching their undisturbed ACT\,+\,Rewind-IL performance (\cref{tab:success_real}). This demonstrates that Rewind-IL handles externally-induced failures as effectively as naturally-occurring ones: the checkpoint database provides a grounded recovery state regardless of cause (\cref{fig:camera_roll_pencil_cup}). The exception is Folding Towel: deformable-cloth manipulation is inherently more variable and multimodal, so the policy's overlapping chunk predictions are intrinsically less self-consistent even under nominal execution, narrowing TIDE's margin between benign variation and genuine failure; more demonstrations could reduce this variability.

\vspace{0.02em}\noindent\textbf{Repeated disturbance resilience:} The +\,Perturb\ comparison applies a single disturbance per rollout, since methods without recovery fail on the first intervention. Rewind-IL places no limit on recovery cycles: because the slot tracker follows the current scene's similarity profile and releases a peaked slot when a newer one surpasses it, successive disturbances that roll the scene back to earlier states naturally re-select the most appropriate VLM-verified checkpoint, not a fixed one. Across the six real tasks, recovery triggers on average $1.26$ times per successful episode, each a clean, independent retry, so retries do not compound.

\begin{table}[t]
\centering
\vspace{-0.5em}
\caption{RoboCasa results: success (\%, 50 rollouts) and detection Bal.\ Acc.\ ($\uparrow$).}
\label{tab:success_sim}
\scriptsize
\renewcommand{\arraystretch}{0.95}
\setlength{\tabcolsep}{4pt}
\setlength{\abovecaptionskip}{2pt}
\setlength{\belowcaptionskip}{0pt}
\begin{adjustbox}{max width=\columnwidth}
\begin{tabular}{lccccc}
\toprule
\multirow{2}{*}{\textbf{Task}} &
\multicolumn{2}{c}{\textbf{Success (\%)}} &
\multicolumn{3}{c}{\textbf{Detection (Bal.\ Acc.)}} \\
\cmidrule(lr){2-3}\cmidrule(lr){4-6}
& ACT & ACT+Rewind-IL & FAIL-Detect & RND & TIDE \\
\midrule
Close Toaster Oven Door & 55 & \textbf{70} & 0.88 & \textbf{0.90} & \textbf{0.90} \\
Open Stand Mixer Head   & 60 & \textbf{80} & 0.88 & 0.89 & \textbf{0.94} \\
Close Fridge            & 55 & \textbf{70} & 0.74 & 0.76 & \textbf{0.94} \\
\midrule
\textbf{Average}        & 56.7 & \textbf{73.3} & 0.83 & 0.85 & \textbf{0.93} \\
\bottomrule
\end{tabular}
\end{adjustbox}
\vspace{-0.5em}
\end{table}

\vspace{0.02em}\noindent\textbf{Disentangling detection and recovery:} Alongside the full system (TIDE\,+\,Checkpoint) we run two conditions, each varying one factor (FD+Ck and TIDE+Sig columns of \cref{tab:success_real}): swapping the detector (FAIL-Detect\,+\,checkpoint) and swapping the recovery (TIDE\,+\,signal-based rewind, which stops at signal clearance, not the verified checkpoint). The recovery axis dominates: replacing checkpoint rewind by signal-based rewind collapses perturbed success to the no-recovery floor (\cref{tab:success_real}), because it stops on the failure-producing trajectory itself. The detection axis is secondary but visible: replacing TIDE by FAIL-Detect costs about $9$\,pp under perturbation, as it both misses more failures and over-triggers during nominal execution. End-to-end success is thus bottlenecked first by \emph{where} it recovers to, then by \emph{when} it triggers.

\begin{wraptable}{r}{0.48\columnwidth}

\centering
\caption{Failure handling computation overhead (10 samples, CUDA).}
\label{tab:timing}
\scriptsize
\renewcommand{\arraystretch}{1.1}
\setlength{\tabcolsep}{2pt}
\begin{tabular}{lc}
\toprule
\textbf{Stage} & \textbf{Mean $\pm$ Std} \\
\midrule
TIDE Computation    & $3.7 \pm 1.0$ \\
Cos-sim Computation\,\,\,\,\,\,\, & $9.6 \pm 0.7$ \\
Slot Bookkeeping    & $5.6 \pm 0.8$ \\
\midrule
\textbf{Total}      & $22.0 \pm 7.2$ \\
\bottomrule
\end{tabular}
\end{wraptable}

\vspace{0.02em}\noindent\textbf{Simulation coverage and detection:} We screen RoboCasa tasks by base-policy competence, since recovery only helps where the base policy can complete the task from a clean state; we retain the three tasks with non-trivial ACT baselines and exclude those ACT fails to acquire (e.g., Coffee Serve Mug, $6\%$). One retained task, \emph{Close Fridge}, is multimodal: its demonstrations differ in door-closing order, inducing a multimodal action distribution, not mere visual variation. Rewind-IL still improves it (\cref{tab:success_sim}), and here the KDE template matters (\cref{sec:Methods}). Simulation failure detection (\cref{tab:success_sim}, detection columns) mirrors the real-robot finding: TIDE attains the highest balanced accuracy, the gain from TNR where distillation baselines over-trigger on multimodal Close Fridge.

\noindent\textbf{Computational overhead:} Per-stage mean and std over 10 CUDA runs appear in \cref{tab:timing}. The three stages (TIDE scoring, cosine-similarity matching, slot bookkeeping) total under $0.2$\,ms, less than $1\%$ of ACT inference time, confirming real-time viability.

\subsection{Adaptation to Flow-Matching Robot Policies}
\vspace{0.02em} \noindent\textbf{Experimental setup:} We replace the CVAE decoder in ACT with a flow-matching (FM) action head~\cite{lipman_flow_2022,black__0_2026} while keeping the transformer encoder architecture identical; TIDE monitors the FM policy's action chunk predictions without any other modification. Detection and task-success results appear in the FM parts of \cref{tab:failure_detection_full} and \cref{tab:success_real}, respectively.

\noindent\textbf{Detection quality transfers to FM:} TIDE again attains the highest balanced accuracy on the FM tasks (\cref{tab:failure_detection_full}, FM rows), staying near-perfect on most and clearly ahead of every baseline. Where FAIL-Detect underperforms on Toolbox\,\&\,Knife, TIDE still achieves perfect detection. TIDE's self-consistency signal thus transfers well to stochastic flow-matching architectures; FM failures often produce even sharper TIDE peaks than ACT, easing threshold calibration.

\noindent\textbf{Rewind-IL delivers consistent gains:} FM\,+\,Rewind-IL improves over the FM baseline on all six real tasks, both undisturbed and under disturbance (\cref{tab:success_real}, FM columns), reaching $90\%$ undisturbed on Cup\,\&\,Box and Pencil\,\&\,Notebook, matching the best ACT condition. This shows the recovery benefit transfers to FM without modification, with gains mirroring the ACT case.

\noindent\textbf{Robustness under disturbance:} Under adversarial disturbance, FM\,+\,Rewind-IL recovers most of the induced loss on the rigid-object tasks but leaves a larger gap on Pencil\,\&\,Notebook and the deformable Folding Towel (\cref{tab:success_real}). Since the FM policy's executed actions stay close to ACT's (with slightly more jitter, an expected effect of its iterative, sampling-based ODE decoding versus ACT's single deterministic pass), its recovery behaves much like ACT's rather than adding margin, so these residual gaps mainly reflect task precision and deformability. Overall, Rewind-IL's core recovery benefit holds broadly across the tested architectures.

\section{Conclusions and Future Work}

We introduce Rewind-IL, an online safeguard framework designed for zero-shot failure detection and recovery for action-chunked imitation learning policies. This is enabled by leveraging temporal inter-chunk discrepancy estimates as a lightweight failure signal and combining it with semantically grounded checkpoint selection for recovery. Avenues for future research include (i) fusing TIDE with complementary signals to improve detection accuracy when policy-intrinsic biases suppress the discrepancy signal; (ii) dynamic slot-count mechanisms for variable-subtask settings; (iii) scene-state verification before committing to recovery.
\bibliographystyle{ieeetr} 
\bibliography{bib}
\end{document}